\documentclass[conference]{IEEEtran}
\usepackage{graphicx}
\usepackage{multirow}
\usepackage{amsmath}
\usepackage{enumitem}
\usepackage[hidelinks]{hyperref}
\usepackage{url}
\usepackage{booktabs}
\usepackage{tabularx}
\usepackage{array}
\usepackage{ragged2e}
\usepackage[caption=false,font=footnotesize]{subfig}
\usepackage{orcidlink}

\title{Beyond the Leaderboard: Design Lessons for Trustworthy Multimodal VQA}

\author{
\IEEEauthorblockN{
Sushant Gautam\orcidlink{0000-0001-9232-2661}\IEEEauthorrefmark{1}\IEEEauthorrefmark{3},
Vajira Thambawita\orcidlink{0000-0001-6026-0929}\IEEEauthorrefmark{1},
Michael A. Riegler\orcidlink{0000-0002-3153-2064}\IEEEauthorrefmark{2},
P{\aa}l Halvorsen\orcidlink{0000-0003-2073-7029}\IEEEauthorrefmark{1}\IEEEauthorrefmark{3},
Steven A. Hicks\orcidlink{0000-0002-3332-1201}\IEEEauthorrefmark{1}
}

\IEEEauthorblockA{
\IEEEauthorrefmark{1}SimulaMet, Norway \qquad
\IEEEauthorrefmark{2}Simula Research Laboratory, Norway \qquad
\IEEEauthorrefmark{3}Oslo Metropolitan University, Norway
}

\IEEEauthorblockA{\{sushant, vajira, michael, paalh, steven\}@simula.no}
}

\begin{document}
\maketitle

\bstctlcite{IEEEexample:BSTcontrol}

\begin{abstract}
Healthcare multimodal AI must combine visual and textual evidence while remaining reliable and interpretable. Using MediaEval Medico 2025 as a retrospective GI endoscopy case study, we analyze design choices across nine documented systems for question answering and explanation quality. Parameter-efficient adaptation of pretrained backbones provides strong challenge performance, but answer-level gains do not consistently translate into faithful and complete clinical reasoning. Methods enforcing structured reasoning and explicit grounding show more reliable behavior across heterogeneous question types, although the evidence is correlational rather than ablation-based. These results motivate evaluation beyond lexical overlap, standardized evidence-linked explanations, leakage-aware data governance, and lightweight robustness and calibration checks. The findings support trustworthy multimodal healthcare AI based on data fusion, explainability, and resilient evaluation.
\end{abstract}


\section{Introduction}

Healthcare AI is increasingly multimodal: clinically useful systems integrate medical images, textual context, and structured decision signals rather than optimize a single modality.

Medical Visual Question Answering (VQA) for gastrointestinal (GI) endoscopy is a concrete setting for this agenda. MediaEval Medico 2025 couples image-grounded question answering with explanation generation on Kvasir-VQA-x1 \cite{medico2025task,gautam2025kvasirx1}, enabling analysis of predictive accuracy and reasoning quality in one pipeline. Instead of treating the benchmark only as a ranking task, we use it as an empirical testbed for multimodal design choices that matter for deployment.

A practical characteristic of this setting is methodological convergence: most systems adapt strong pretrained vision-language models using parameter-efficient fine-tuning (PEFT), typically LoRA or QLoRA~\cite{hu2022lora,QLORA,PEFT}. While not a controlled ablation and lacking zero-shot frontier VLM baselines, this enables descriptive comparison of structured reasoning, evidence grounding, confidence reporting, and robustness handling.

Our contributions in this paper are:
\begin{itemize}[leftmargin=*]
  \item \textbf{Multimodal data integration and diagnosis support:} a cross-team taxonomy of image-language adaptation, reasoning, and grounding strategies for GI decision support.
  \item \textbf{Explainable and resilient AI:} evidence-driven findings on recurring failure modes tied to faithfulness, completeness, and robustness across question categories and complexity levels.
  \item \textbf{Performance evaluation methodology:} practical recommendations for trustworthy multimodal evaluation, including semantic correctness checks, standardized explanation artifacts, calibration metadata, and leakage-explicit reporting.
\end{itemize}

These findings guide multimodal medical imaging, NLP-integrated clinical decision support, and reliable real-world healthcare AI.

\section{Related Work}

Medical Visual Question Answering (MedVQA) has evolved from early discriminative systems toward generative Vision-Language Models (VLMs) capable of open-ended clinical reasoning~\cite{liu2025medicalvlm}. Initial shared tasks such as VQA-Med emphasized classification-style answering, where models selected responses from constrained vocabularies~\cite{ImageCLEFVQA-Med2019}. While suitable for controlled evaluation, such approaches limited the expressive capacity required for real-world clinical reasoning, where nuanced natural language generation is often necessary~\cite{liu2025medicalvlm, ImageCLEFVQA-Med2019}.
Subsequent surveys of MedVQA highlight recurring structural challenges: limited dataset scale, cross-institutional domain shift, and the gap between fluent response generation and clinically trustworthy reasoning~\cite{bansal2023medical,Lin2023Sep,ImageCLEFVQA-Med2019}.

\subsection{Endoscopic VQA and Kvasir Dataset Lineage}

GI endoscopy introduces modality-specific complexity, including specular reflections, occlusion by instruments, non-standard orientations, and fine-grained localization language~\cite{ImageCLEFmedicalVQA}. The Kvasir-VQA dataset was an early effort to adapt GI imaging to the VQA paradigm, enabling attribute-based reasoning over polyps, landmarks, and procedural artifacts~\cite{Kvasir-VQA}.
More recent iterations, including Kvasir-VQA-x1 \cite{Kvasir-VQA,gautam2025kvasirx1}, scale both the number of question–answer pairs and the compositional reasoning depth, introducing structured multi-hop prompts that require synthesizing count, color, location, and presence cues within a single response.
This progression reflects a broader shift in MedVQA toward reasoning-centric benchmarks rather than purely lexical matching tasks.

\subsection{Parameter-Efficient Adaptation of Large Models}

As large VLMs became dominant, research focus moved from designing task-specific architectures to adapting pretrained multimodal backbones~\cite{PEFT}. In clinical contexts, full-parameter fine-tuning of large language models is often computationally impractical~\cite{liu2025medicalvlm}. PEFT methods, particularly LoRA, enable domain adaptation by training small rank-decomposed update matrices while freezing the base model~\cite{hu2022lora}.
QLoRA~\cite{QLORA} further reduces memory footprint, allowing adaptation of billion-parameter models under modest hardware constraints.

Recent surveys of medical Vision Language Models (VLM) emphasize that while such adaptation improves answer quality, it does not inherently guarantee clinically faithful reasoning, reinforcing the need for structured explanation and grounding strategies~\cite{liu2025medicalvlm}.
Studies on LoRA and PEFT~\cite{mmssnce2025,aeolus2025,PEFT} further demonstrate the practical impact of parameter-efficient adaptation in medical VQA.

\subsection{Explainability, Faithfulness, and Clinical Trust}

The `black box' nature of deep generative models raises particular concern in medicine, where incorrect but confidently stated answers can have significant consequences~\cite{liu2025medicalvlm}. Early work in clinically applicable deep learning demonstrated strong diagnostic performance while simultaneously highlighting the importance of interpretable evidence for clinical adoption.
In the VQA domain, the VQA-X dataset formalized explanation generation as a supervised objective, introducing textual and visual rationales as first-class outputs~\cite{vqa-x}.
Subsequent research distinguishes between \emph{plausibility} (whether explanations sound reasonable) and \emph{faithfulness} (whether they reflect the model’s actual decision process). Faithfulness-oriented frameworks advocate perturbation testing, consistency checks, and structured evaluation criteria rather than relying solely on fluency~\cite{jacovi2020faithfulness}.

Recent surveys further formalize faithfulness and completeness as core axes of explainable AI evaluation, arguing that explanation quality must be measured independently from answer correctness~\cite{Lyu2024Jun}.
Parallel developments in evaluation methodology introduce rubric-based large language model (LLM) adjudication for scalable multi-dimensional grading. While enabling structured assessment of clarity, faithfulness, and completeness, LLM-as-judge frameworks also introduce judge-model dependence and prompt sensitivity~\cite{Li2024Dec}.

Collectively, these strands of research motivate benchmarks that assess not only answer generation but also grounded, faithful, and clinically interpretable reasoning under realistic computational constraints.

\section{Problem Setup and Study Design}

\begin{figure}[!t]
\centering
\begin{minipage}{0.9\linewidth}
\centering
\scriptsize
\begin{minipage}[t]{0.46\linewidth}
\centering
\textbf{VQA (Task 1) }\\
\vspace{1mm}
\includegraphics[width=0.8\linewidth]{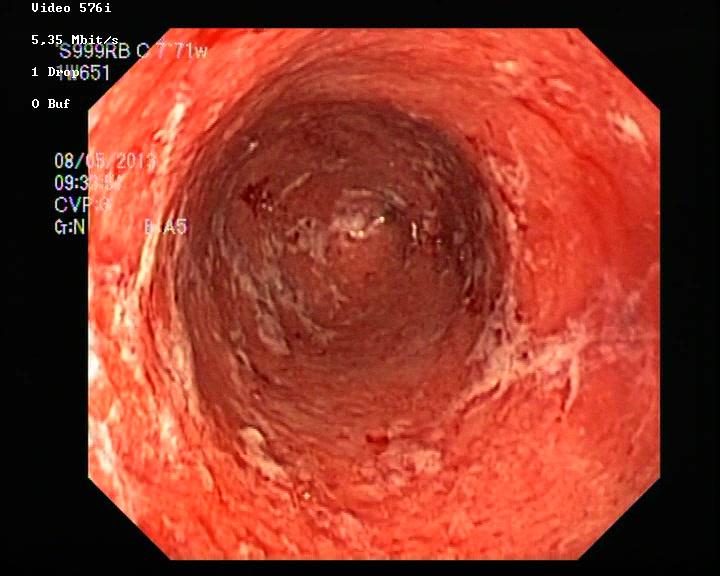}\\
\raggedright
\textbf{Question:} Is esophagitis visible?\\
\vspace{1mm}
\textbf{Model answer:} Yes, mild distal esophagitis is present.
\end{minipage}
\begin{minipage}[t]{0.53\linewidth}
\centering
\textbf{Explainable Reasoning (Task 2)}\\
\raggedright
Probe 1: \textit{Where is the finding?}\\
Answer 1: \textit{...}\\

Probe 2: \textit{What color is the finding?}\\
Answer 2: \textit{...}\\

\textit{...}\\

Probe 10: \textit{How many ... ?}\\
Answer 10: \textit{...}\\
\vspace{1mm}
\textbf{Explanation} (LLM-summarized): 
\textit{Model is confident that image indeed contains esophagitis because... }
\end{minipage}

\end{minipage}

\caption{Example of Task 1 (visual question answering) followed by probing-guided explanation~\cite{teamnepal2025} for explainable multimodal reasoning for Task 2.}
\label{fig:task_description}
\end{figure}

We use Medico 2025 as a study setting with two connected subtasks: answer generation for GI image-question pairs in Kvasir-VQA-x1 and multimodal explanation aligned with the predicted answer, emphasizing clinical usefulness, faithfulness, and interpretability \cite{medico2025task}.
This paper performs a post-hoc, cross-system analysis. In scope are: (i) official Public and Final leaderboard outcomes, (ii) organizer-released stratified post-challenge diagnostics, and (iii) method characterization from working notes.
We do not treat self-reported internal metrics as a single comparable benchmark, because teams used different validation subsets, split strategies, and reporting conventions.
Figure~\ref{fig:task_description} shows a qualitative task-flow example that combines a Task 1 image-question-answer instance with probing-based explanation generation.

The task uses de-identified gastrointestinal (GI) imaging data and associated question-answer annotations without personal identifiers or patient-specific information.

\section{Dataset}

Kvasir-VQA-x1~\cite{gautam2025kvasirx1} extends the original Kvasir-VQA~\cite{Kvasir-VQA} benchmark into a large-scale multimodal reasoning resource for GI endoscopy. It comprises 6,500 source images and 159,549 question–answer (QA) pairs, each annotated by question class and compositional complexity \cite{medico2025task}. Complexity is defined by reasoning depth: Level~1 (single atomic QA), Level~2 (two merged atomic QAs), and Level~3 (three merged atomic QAs), with an approximately balanced distribution across levels. As described in the challenge, atomic QA pairs are systematically merged into single-, two-, and three-hop prompts and rewritten into clinical language using LLM support \cite{medico2025task}, following an image-centric construction strategy aligned with the intended complexity formulation.

Question coverage includes yes/no, choice-based, color, location, and numeric count forms, with clinically oriented classes such as polyp type, instrument presence, landmark localization, and finding count. This mix creates a realistic tension between lexical matching and true visual grounding: many questions are short and templated, while others require multi-attribute synthesis.
From a modeling perspective, likely hard cases reflect endoscopy-specific nuisances and structured reasoning demands: specular highlights, low contrast, blur, occlusion by tools, and ambiguity in localization language.

A key caveat for result interpretation is split hygiene. Team Lama4Vision reports substantial image-level overlap between train and test partitions under QA-level splitting and recommends strict image-level partitioning in future releases \cite{lama2025}. Given the image-centric QA generation process, this risk is structurally plausible, although no complete per-team overlap audit was available.
Importantly, image identifiers were available; thus, participants could, in principle, construct image-level disjoint training and validation splits, even if the official QA-level split contained overlapping images.
To further assess robustness and mitigate leakage concerns, a separate private set, similar to Kvasir-VQA-x1 and derived from the ImageCLEFmed MEDVQA 2025 challenge set, was used~\cite{ImageCLEFmedicalVQA}. It contains 500 images not present in Kvasir-VQA-x1 and 5,368 question–answer pairs across the three defined complexity levels, enabling a cleaner cross-dataset generalization assessment.
Thus, overlap is treated here as a risk requiring cautious interpretation rather than proof of contamination, motivating leakage-explicit evaluation protocols in future MedVQA benchmarks.

\section{Evaluation Protocol}

We use two ranking labels throughout: \emph{Public Leaderboard}, which refers to the submission-time leaderboard feedback on a 1,500-question subset of the Kvasir-VQA-x1 test set (Subtask 1 only); and \emph{Final Leaderboard}, which denotes the organizer-reported final ranking table after organizer-side evaluation on full test set and private set (used for Subtask 1 and Subtask 2 ranking claims).

\subsection{Subtask 1: Class-wise Semantic Adjudication}
In addition to lexical summaries, the official post-challenge Subtask 1 release includes per-item semantic adjudication statistics for both test and private set. For each generated answer, Qwen3-30B-A3B\footnote{Checkpoint: \url{https://huggingface.co/Qwen/Qwen3-30B-A3B}.} \cite{qwen3technicalreport} evaluates aligned prediction-question-gold tuples (with complexity and source atomic-QA context) and assigns binary labels (1/0) for required question-class aspects, plus short rationales. Aggregating these aspect labels yields category- and complexity-wise diagnostic profiles.

\subsection{Subtask 2: Rubric-Based LLM Adjudication}
Subtask 2 evaluation also uses Qwen3-30B-A3B as a rubric-based adjudicator and scores each item on five dimensions: Answer Correctness, Faithfulness, Clinical Relevance, Clarity, and Completeness \cite{medico2025results,qwen3technicalreport}. This reflects the final scoring procedure used for ranking, replacing earlier expert evaluation components proposed in \cite{medico2025task} that were not executed at scale. Scores are floating-point values between 0 and 1, with per-dimension rationales. Subtask 2 ranking is reported via the Final Leaderboard (there is no Subtask 2 Public Leaderboard).

\subsection{Evaluation Protocol Used in This Analysis}
To interpret behavior beyond headline ranks, we use official evaluation artifacts together with team working notes. For Subtask 1, we use lexical summaries of the full test set and private set to report ROUGE-1/2/L \cite{lin2004rouge}, METEOR \cite{banerjee2005meteor}, chrF++ \cite{popovic2017chrfpp}, BLEU \cite{papineni2002bleu}, and BERTScore \cite{zhang2020bertscore} trends~\cite{evaluate2022}, plus class-wise semantic adjudication outputs for category-level strengths and weaknesses. For Subtask 2, we use rubric scores values between 0 and 1 and question-type aggregates for the five explanation dimensions.
Team working notes are used only for qualitative mechanism interpretation, not for re-ranking. Ranking claims are grounded in Final Leaderboard values; full test set and private set outputs are used for behavior analysis.

\subsection{Threats to Validity}
Key risks are lexical bias in Subtask 1 text-overlap metrics, judge-model and prompt dependence in semantic scoring, limited direct verification of visual evidence in text-centric diagnostics, and non-uniform reporting across working notes. Qwen3-30B-A3B lacked clinician validation at scale here and may introduce alignment bias if submitted systems use related model families. We mitigate these risks by anchoring ranking statements to Final Leaderboard values, labeling post-challenge diagnostics, and emphasizing cross-paper consensus over isolated claims. LLM-judged diagnostics remain structured proxies rather than substitutes for clinician-led assessment.

\section{Compared Methods and Design Axes}

Nine registered working-note papers cover a diverse but convergent method space \cite{teamnepal2025,mmssnce2025,cvgiba2025,endovision2025,selab2025,aeolus2025,lama2025,ssn2025,irel2025}. These papers reflect the submissions that were formally documented; not all participating teams, or submitted systems, ultimately resulted in a working-note paper. Across the reported approaches, most systems start from a general-purpose VLM and apply parameter-efficient adaptation, with variation concentrated in explanation generation, visual grounding, and robustness handling. The most common backbones include Florence-2~\cite{xiao2023florence2advancingunifiedrepresentation}, PaliGemma~\cite{beyer2024paligemmaversatile3bvlm,steiner2024paligemma2familyversatile}, BLIP-2~\cite{li2023blip2bootstrappinglanguageimagepretraining}, InstructBLIP~\cite{dai2023instructblipgeneralpurposevisionlanguagemodels}, and Qwen2-VL~\cite{wang2024qwen2vlenhancingvisionlanguagemodels}, while CLIPSeg~\cite{CLIPSeg} appears as a grounding module. Methods such as Grad-CAM~\cite{Selvaraju}, attention rollout~\cite{AttentionRollout}, and curriculum learning~\cite{Soviany2021Jan} also appear in team reports.

For quantitative diagnostics, we use subtask evaluation outputs across submitted systems: lexical stratified summaries and class-wise semantic adjudication for Subtask 1, plus rubric-based semantic outputs for Subtask 2. This supports a design-axis analysis while preserving Final Leaderboard tables as the primary ranking reference.
Table~\ref{tab:taxonomy} summarizes the cross-team method taxonomy used in this study.

\begin{table*}[!t]
\caption{Cross-team method taxonomy from Medico 2025 working notes.}
\label{tab:taxonomy}
\centering

\setlength{\tabcolsep}{3pt}        
\begin{tabularx}{\textwidth}{@{}
>{\RaggedRight\arraybackslash}p{2.2cm}
>{\RaggedRight\arraybackslash}p{2.5cm}
>{\RaggedRight\arraybackslash}p{2.5cm}
>{\RaggedRight\arraybackslash}p{5.5cm}
>{\RaggedRight\arraybackslash}p{4.4cm}
@{}}
\toprule
\textbf{Family} & \textbf{Teams} & \textbf{Backbone(s)} & \textbf{Explanation and grounding strategy} & \textbf{Efficiency and robustness choices} \\
\midrule
Self-probing pipelines
& Team Nepal~\cite{teamnepal2025}; IReL@IIT(BHU)~\cite{irel2025}
& PaliGemma-3B; Florence~-2 (IReL dual-track)
& Auxiliary clinical sub-questions before final narrative; Team Nepal adds LLM synthesis, while IReL reports token-probability confidence
& QLoRA/LoRA PEFT~\cite{hu2022lora,QLORA,PEFT}; IReL adds specular-inpainting preprocessing for Florence-2~\cite{xiao2023florence2advancingunifiedrepresentation} \\
\midrule
Multi-task grounded learning
& CVG-IBA~\cite{cvgiba2025}
& Florence-2
& Joint training for answering, explanation text, and text-to-region grounding; CLIPSeg~\cite{CLIPSeg}/Kvasir-SEG supervision; task tokens for inference modes
& High-rank LoRA; shared multi-task training as regularization \\
\midrule
Joint answer-plus-explanation heads
& SSN-InnovateX~\cite{ssn2025}
& BLIP-2
& Unified output with textual rationale, attention-rollout maps~\cite{AttentionRollout}, and confidence estimation
& Single-backbone joint training for answer and explanation generation \\
\midrule
Strong answering baselines
& MM-SSNCE~\cite{mmssnce2025}; Aeolus~\cite{aeolus2025}; EndoVision~\cite{endovision2025}
& PaliGemma-3B; PaliGemma~2
& Primarily answer-focused training; Aeolus/EndoVision report Grad-CAM~\cite{Selvaraju}-based visual evidence
& LoRA-based PEFT and quantization for MM-SSNCE/Aeolus; EndoVision reports a lightweight explanation module and confidence cues \\
\midrule
Alternative adaptation strategies
& SELab-HCMUS~\cite{selab2025}; Lama4Vision~\cite{lama2025}
& Qwen2-VL-2B~\cite{wang2024qwen2vlenhancingvisionlanguagemodels}; InstructBLIP (Flan-T5-XXL)~\cite{dai2023instructblipgeneralpurposevisionlanguagemodels}
& Limited explicit grounding; focus on answer quality and complexity-aware behavior
& Encoder-decoder LoRA; curriculum-guided LoRA~\cite{Soviany2021Jan} with weak augmentation and leakage analysis \\
\bottomrule
\end{tabularx}
\end{table*}

Two practical trends stand out. First, parameter-efficient tuning is the dominant strategy: most teams rely on LoRA/QLoRA variants~\cite{hu2022lora,QLORA,PEFT} rather than full retraining at scale. Second, explanation quality is often treated as a pipeline design problem, with explicit reasoning structure (self-probing or multi-task grounding) used as a recurring design pattern.

Overall novelty is incremental but meaningful: the benchmark did not produce radically new architectures, but it clarified which lightweight additions were repeatedly adopted under practical compute constraints.

\section{Results and Analysis}

\begin{table*}[!t]
\caption{Performance of the top four ranked teams in MediaEval Medico 2025 for each subtask. Subtask 1 reports private-set lexical scores with Test set$\rightarrow$Private set BLEU drift; Subtask 2 reports rubric-based explanation scores.}
\label{tab:results}
\centering
\small
\setlength{\tabcolsep}{4pt} 

\subfloat[Task 1: GI Visual Question Answering (on Private Set)]{
\label{tab:results_a}
\begin{minipage}[t]{0.49\linewidth}
\centering
\resizebox{\linewidth}{!}{
\begin{tabular}{lcccc}
\toprule
\textbf{Team} & \textbf{BLEU} & \textbf{ROUGE-L} & \textbf{METEOR} & \textbf{$\Delta$BLEU} \\
\midrule
MM-SSNCE~\cite{mmssnce2025} & 0.47 & 0.67 & 0.67 & $-0.023$ \\
Team Nepal~\cite{teamnepal2025} & 0.47 & 0.67 & 0.67 & $-0.023$ \\
CVG-IBA~\cite{cvgiba2025} & 0.45 & 0.65 & 0.65 & $-0.048$ \\
EndoVision~\cite{endovision2025} & 0.42 & 0.64 & 0.64 & $-0.005$ \\
\bottomrule
\end{tabular}
}
\end{minipage}
}
\hfill
\subfloat[Task 2: Explainable Multimodal Reasoning]{
\label{tab:results_b}
\begin{minipage}[t]{0.48\linewidth}
\centering
\resizebox{\linewidth}{!}{
\begin{tabular}{lccccc}
\toprule
\textbf{Team} & \textbf{Corr.} & \textbf{Faith.} & \textbf{Clin. Rel.} & \textbf{Clarity} & \textbf{Comp.} \\
\midrule
Team Nepal~\cite{teamnepal2025} & 0.86 & 0.74 & 0.76 & 0.90 & 0.61 \\
CVG-IBA~\cite{cvgiba2025} & 0.74 & 0.52 & 0.57 & 0.82 & 0.45 \\
SSN-InnovateX~\cite{ssn2025} & 0.74 & 0.57 & 0.58 & 0.63 & 0.46 \\
IReL@IIT(BHU)~\cite{irel2025} & 0.59 & 0.27 & 0.38 & 0.82 & 0.25 \\
\bottomrule
\end{tabular}
}
\end{minipage}
}

\end{table*}

\begin{figure*}[!t]
\centering
\subfloat[Subtask 1 results across question classes and complexity levels. Plot highlights non-monotonic difficulty across complexity and category.]{%
\label{fig:organizer_shared_diagnostics_a}
\includegraphics[width=0.49\textwidth]{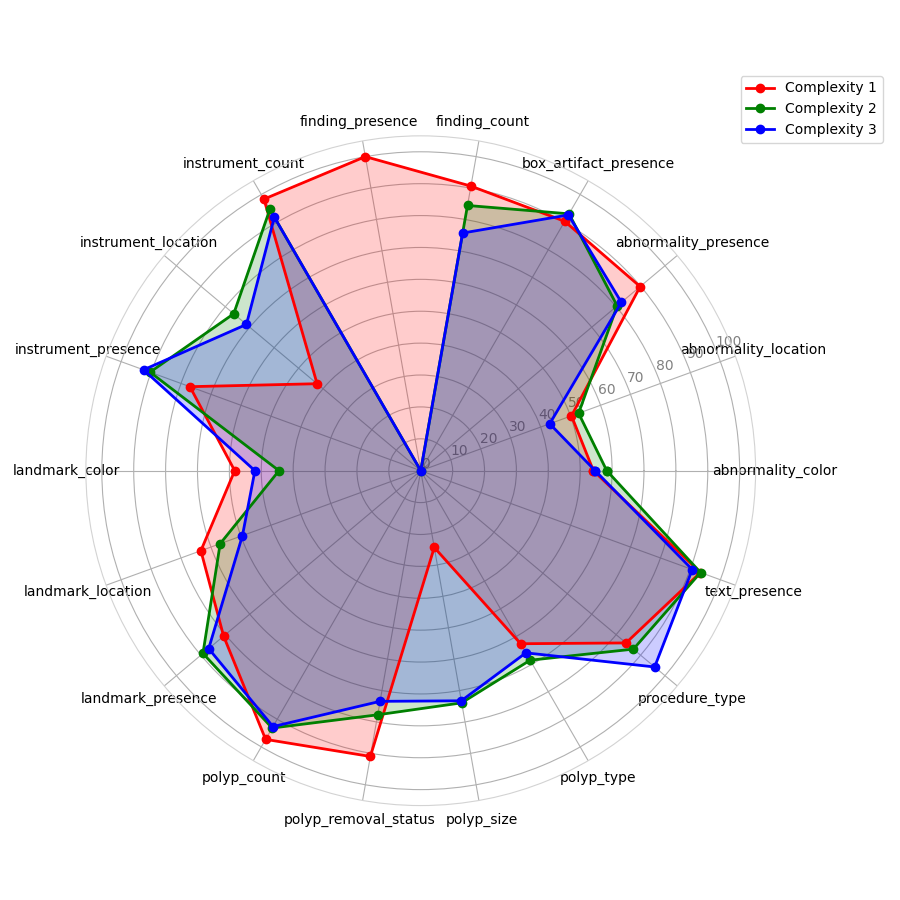}%
}
\hfill
\subfloat[Subtask 2 score distribution by question types. Distribution shows that clarity can remain high while faithfulness and completeness vary on harder question types]{%
\label{fig:organizer_shared_diagnostics_b}
\includegraphics[width=0.49\textwidth]{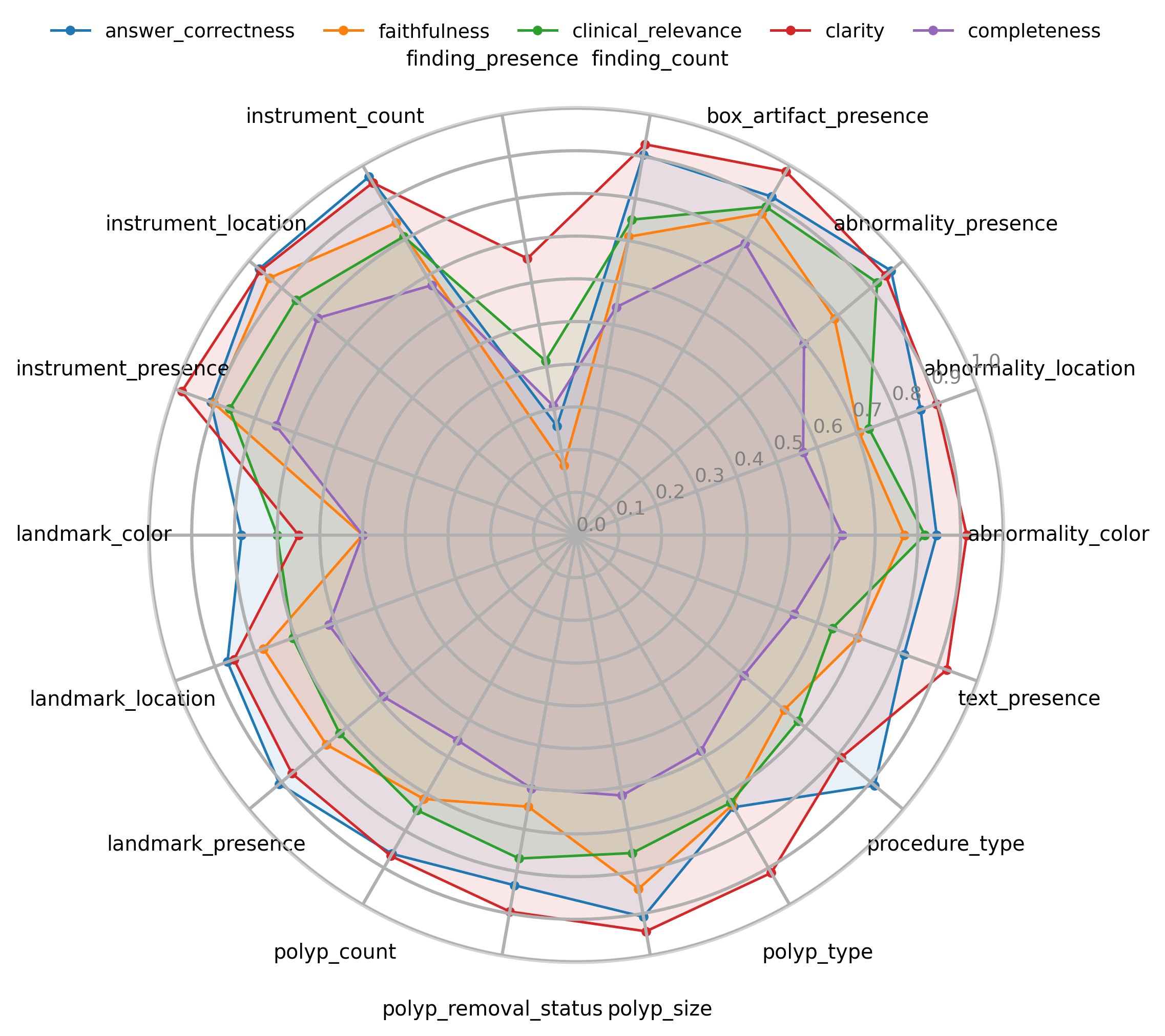}%
}
\caption{Diagnostics plots for Subtask 1 and Subtask 2 (from Team Nepal).}
\label{fig:organizer_shared_diagnostics}
\end{figure*}

Table~\ref{tab:results}\subref{tab:results_a} summarizes Task 1 leaderboard metrics, while Table~\ref{tab:results}\subref{tab:results_b} summarizes Task 2 rubric scores. We organize the analysis around four questions:

\subsection{RQ1: Are headline leaderboard gaps practically meaningful?}

Task 1 headline standings depend on which split is examined. In test set evaluation, CVG-IBA has the highest BLEU (0.502), while in the Private set MM-SSNCE and Team Nepal tie for first (0.471). Thus, rank order is not perfectly stable across splits even when absolute gaps are modest.

Across all nine teams with both splits, the mean absolute Test set $\leftrightarrow$Private set BLEU drift is 0.021, and all teams remain within $\pm 0.05$ BLEU. This indicates moderate stability overall.

The test score distribution across all nine teams also supports a ``tight top tier'' interpretation, but only for the leading cluster. For BLEU, the top quartile starts at 0.494 (Q3), with a narrow gap to the maximum of 0.502; analogous Q3-to-max gaps are similarly small for ROUGE-L (0.697 to 0.705) and METEOR (0.701 to 0.704). The broader field remains more dispersed (e.g., BLEU median 0.427, minimum 0.154), indicating simultaneous top-tier clustering and long-tail spread.

\subsection{RQ2: Where do errors concentrate across complexity and category?}

The complexity profile is clearly non-monotonic at the benchmark level. Averaged across all nine teams on the test split, BLEU is 0.356 (Complexity 1; L1), 0.323 (Complexity 2; L2), and 0.416 (Complexity 3; L3). ROUGE-L follows the same pattern: 0.624, 0.582, and 0.612. Complexity-2 appears consistently hardest on lexical metrics, while Complexity-3 often rebounds.
Team-level behavior supports this: 6/9 teams show non-monotonic BLEU trajectories across complexities (i.e., neither strictly increasing nor strictly decreasing), reinforcing that question composition matters more than nominal difficulty labels alone.

Class-wise semantic adjudication (Fig.~\ref{fig:organizer_shared_diagnostics}\subref{fig:organizer_shared_diagnostics_a}) shows persistent weak classes. Lowest mean class scores are \texttt{landmark\_color} (0.494), \texttt{abnormality\_location} (0.500), \texttt{polyp\_size} (0.508), and \texttt{instrument\_location} (0.539). Strongest classes are \texttt{polyp\_count} (0.935), \texttt{finding\_presence} (0.929), \texttt{instrument\_count} (0.915), and \texttt{box\_artifact\_presence} (0.877).
Fig.~\ref{fig:organizer_shared_diagnostics}\subref{fig:organizer_shared_diagnostics_a} is consistent with this pattern: classes requiring fine spatial/color discrimination remain weaker than counting/presence-oriented classes.

A complexity-by-class analysis shows interaction effects that are not visible from class-only averages. For example, the mean for \texttt{instrument\_location} across teams rises from 0.306 at L1 to 0.705 at L2, and \texttt{polyp\_size} rises from 0.191 (L1) to about 0.658 (L2/L3), suggesting some classes are especially difficult only at specific levels. Conversely, \texttt{landmark\_color} remains weak at L2/L3 (0.420/0.426), indicating a more persistent color-understanding bottleneck.

\subsection{RQ3: How robust are systems under split shift and artifacts?}

Split robustness is mixed. Most teams exhibit small to moderate Test$\rightarrow$Private drift, but some teams undergo meaningful metric contraction. The spread of BLEU deltas (from about $-0.048$ to $+0.022$) suggests that robustness is method-dependent rather than uniformly achieved.

A ranked drift view clarifies where degradation concentrates: CVG-IBA ($-0.048$) and SELab-HCMUS ($-0.030$) show the largest BLEU drops, while EndoVision ($-0.005$) and IReL ($+0.003$) are nearly unchanged. Three teams improve on Private (Aeolus $+0.022$, Lama4Vision $+0.018$, IReL $+0.003$), indicating that split shift does not uniformly penalize all systems.

Team reports nevertheless converge on practical robustness interventions: light geometric/color augmentation (Team Nepal, Lama4Vision, Aeolus), artifact-aware preprocessing such as specular inpainting (IReL), and curriculum training (Lama4Vision) \cite{teamnepal2025,lama2025,aeolus2025,irel2025}. These are plausible contributors, but current evidence remains correlational because standardized perturbation stress-tests were not part of official reporting.

Split hygiene also affects interpretation. The overlap concern reported by Lama4Vision indicates that some generalization claims may be inflated unless strict image-disjoint protocols are enforced \cite{lama2025}.
At the same time, drift alone is insufficient for strong leakage attribution. The current benchmark metadata includes explicit overlap discussion for Lama4Vision, but not a complete per-team overlap audit. Therefore, split-delta evidence should be interpreted as a risk indicator rather than proof of contamination.

\subsection{RQ4: What determines explanation trustworthiness?}

Subtask 2 variance is driven mainly by grounding dimensions, not fluency alone. Across teams, the widest ranges are faithfulness (0.474), clinical relevance (0.371), and completeness (0.368), while clarity varies less (0.279). This confirms that fluent language is easier to optimize than faithful, clinically grounded justification.

Question-type aggregation sharpens the failure mode. \texttt{finding\_presence} is by far the hardest type (correctness 0.099, faithfulness 0.061, clarity 0.663), giving the largest clarity-minus-faithfulness gap (+0.602). In contrast, \texttt{instrument\_count} and \texttt{polyp\_count} are strongest in correctness (0.878 and 0.868), but \texttt{instrument\_count} still has the largest correctness-minus-completeness gap (+0.432), indicating that short answers can be right while explanations remain shallow.
This pattern is also visible in Fig.~\ref{fig:organizer_shared_diagnostics}\subref{fig:organizer_shared_diagnostics_b}, where clarity remains relatively high while faithfulness and completeness degrade on harder question types.

Taxonomy-level comparison (Table~\ref{tab:taxonomy}) is informative but statistically limited by few teams per family. Hence, family-vs-outcome results should be read as exploratory descriptive evidence rather than causal or inferential proof.

At team level, Team Nepal leads all five dimensions and has the strongest balance between correctness and explanatory grounding \cite{medico2025results}. IReL shows the clearest fluent-but-under-grounded profile (clarity 0.824 vs faithfulness 0.269 and completeness 0.246) \cite{medico2025results,irel2025}.

\subsection{Practical Trade-offs: Efficiency and Explainability}

Efficiency is a strong positive outcome of this benchmark setting. Most teams achieved competitive performance with LoRA/QLoRA adapters and 4-bit quantization rather than full-parameter tuning \cite{teamnepal2025,mmssnce2025,aeolus2025,irel2025}. Reported training hardware includes NVIDIA T4 GPU while still reaching high leaderboard positions. These results reflect what was achieved under practical compute constraints. They do not rule out that larger or more powerful fully fine-tuned models could obtain higher absolute scores given substantially greater resources.

The main practicality trade-off is the cost of explainability. Self-probing pipelines add extra inference steps and synthesis overhead but can improve explanatory structure. Joint multi-task approaches amortize explanation generation inside one backbone, but they require additional supervision artifacts (for example text-mask pairs) and more complex training curation \cite{cvgiba2025,ssn2025}. Subtask 2 results suggest this added cost is justified when it measurably increases faithfulness and completeness.

A remaining gap is standardized reporting: latency, throughput, and calibration quality were not consistently reported across teams. Future MedVQA evaluations should require minimal efficiency metadata to support real deployment decisions.

\section{Discussion: Implications for Multimodal Healthcare AI}
The MediaEval Medico 2025 challenge study indicates a pattern relevant beyond one benchmark: when teams share strong pretrained backbones and limited adaptation budgets, PEFT~\cite{PEFT} becomes the default strategy. Under this convergence, performance differences are driven less by backbone choice and more by how multimodal evidence is integrated and validated. Explicit reasoning structure, grounding constraints, and calibrated confidence reporting are consistently associated with stronger trust signals than prompt-only refinement. LoRA/QLoRA~\cite{hu2022lora,QLORA} with 4-bit quantization~\cite{liu-etal-2023-llm} was common among top-performing systems.

From a medical AI perspective, this supports three broad conclusions. First, multimodal healthcare progress should be evaluated as a data-fusion problem, not only a text-generation problem. Second, explainability must be operationalized through verifiable evidence links, not fluency alone. Third, resilient deployment requires robustness and uncertainty reporting as first-class outputs, especially for noisy, incomplete, or shifted clinical inputs.

This retrospective also has limitations. Although submissions were re-evaluated in a unified environment, part of the design interpretation still depends on participant-reported working notes, and semantic adjudication remains judge-model dependent. Therefore, the strongest claims are comparative and methodology-oriented.

\label{sec:recommendations}
We recommend five practical upgrades for multimodal evaluation and reporting.

\textbf{1) Evaluate multimodal correctness beyond lexical overlap.}
Keep BLEU/ROUGE/METEOR for continuity, but add low-cost supplements: numeric exact-match for counts, yes/no normalization, concept-level matching for clinically equivalent phrasings, and a fixed semantic-adjudication protocol for class-wise correctness.

\textbf{2) Report stratified performance and stress tests as standard artifacts.}
Release per-complexity and per-question-class summaries for original and transformed settings, and tag lexical versus semantic-adjudication tables/plots. Add small corruption and transformation tests to quantify robustness under realistic shifts without large annotation overhead.

\textbf{3) Enforce leakage-explicit and bias-aware data governance.}
Require image-level disjoint train/validation/test partitions (or explicit overlap reporting), plus a short split-audit statement in each submission paper \cite{lama2025}. To reduce organizer burden, expose image identifiers and simple overlap scripts; when metadata permits, include subgroup diagnostics to flag potential bias.

\textbf{4) Standardize explainability outputs and calibration metadata.}
Define one schema for textual explanation, visual evidence object (bbox/mask/heatmap with coordinates), and confidence fields. To limit annotation cost, require four minimal textual elements (\emph{where}, \emph{what}, \emph{how sure}, \emph{what else}), plus a confidence-computation and calibration note.

\textbf{5) Add lightweight resilience and faithfulness checks.}
Introduce probes such as region-masking sensitivity and answer-explanation consistency checks. These checks add inference overhead, but small standardized tests can already discourage fluent but ungrounded narratives.

\section{Conclusion}

Using Medico 2025 as a case study, this paper provides transferable guidance for multimodal AI in healthcare. Its novelty relative to prior MedVQA surveys is quantified cross-system evidence from a real challenge, not a new architecture. Reliable clinical behavior depends more on principled multimodal integration and evaluation than on larger backbones alone: structured reasoning, explicit grounding, and calibrated reporting provide the clearest trust signals. The recommendations in Section~\ref{sec:recommendations} target practical adoption in multimodal AI and deployment.

The findings support healthcare data fusion, multimodal disease diagnosis, NLP-integrated medical imaging, explainable AI, resilient systems, and performance evaluation. Future work should extend this framework to broader modalities (for example EHRs, genomics, and wearables) with stronger fairness and privacy auditing for personalized medicine settings.

\section*{Acknowledgments}

We thank the MediaEval community and all participating teams for open reporting and reproducible challenge contributions. The research presented in this paper has benefited from the Experimental Infrastructure for Exploration of Exascale Computing (eX3), which is financially supported by the Research Council of Norway under contract 270053.

\bibliographystyle{IEEEtran}
\bibliography{references}
\end{document}